\documentclass[fleqn,10pt]{wlscirep}

\usepackage[utf8]{inputenc}
\usepackage[T1]{fontenc}
\usepackage{lineno}
\usepackage{xcolor}

\usepackage{multirow}
\usepackage{makecell}

\title{An annotated grain kernel image database for visual quality inspection}

\author[1,2,$^\dagger$ *]{Lei Fan}
\author[1,*]{Yiwen Ding}
\author[1]{Dongdong Fan}
\author[1]{Yong Wu}
\author[1]{Hongxia Chu}
\author[2]{Maurice Pagnucco}
\author[2,$^\dagger$]{Yang Song}
\affil[1]{Gaozhe Technology, Hefei, 230088, China}
\affil[2]{School of Computer Science and Engineering, Faculty of Engineering, The University of New South Wales, Sydney, 2052, Australia}

\affil[$^\dagger$]{corresponding author(s): Lei Fan (lei.fan1@unsw.edu.au and yang.song1@unsw.edu.au)}
\affil[*]{these authors contributed equally to this work}

\def\etal{\emph{et al. }}
\def\ie{\emph{i.e., }}
\def\eg{\emph{e.g., }}

\begin{abstract}

We present a machine vision-based database named \textit{GrainSet} for the purpose of visual quality inspection of grain kernels. The database contains more than 350$K$ single-kernel images with experts' annotations. The grain kernels used in the study consist of four types of cereal grains including wheat, maize, sorghum and rice, and were collected from over 20 regions in 5 countries. The surface information of each kernel is captured by our custom-built device equipped with high-resolution optic sensor units, and corresponding sampling information and annotations include collection location and time, morphology, physical size, weight, and Damage \& Unsound grain categories provided by senior inspectors. In addition, we employed a commonly used deep learning model to provide classification results as a benchmark. We believe that our \textit{GrainSet} will facilitate future research in fields such as assisting inspectors in grain quality inspections, providing guidance for grain storage and trade, and contributing to applications of smart agriculture. 

\end{abstract}
\begin{document}

\pagestyle{empty}
\pagenumbering{gobble} 
\flushbottom
\maketitle

\section*{Background \& Summary}

Cereal grains play an essential role in the human diet by providing a diverse range of nutrients \cite{mckevith2004nutritional,stevenson2012wheat}, including carbohydrates, dietary fiber and vitamins. Grain kernels are consumed in a variety of food products \cite{charalampopoulos2002application}, such as bread, pasta and cereals, and serve as raw resources for animal feed, ethanol, adhesives, and other necessities \cite{abuajah2015functional,bigliardi2013innovation}. The inspection of grain kernels prior to processing is a critical step in ensuring the quality, safety and efficiency of the production process \cite{evers2002cereal}, while also enhancing consumer health and supporting the economic value of cereal crops \cite{cassman2003meeting}. For example, The inspection is a pre-requisite for preventing the spread of harmful contaminants (\eg mycotoxins \cite{placinta1999review}). 

Several organizations have established guidelines for the inspection of grain quality. For example, the United States Department of Agriculture (USDA) \cite{usda} stipulates standards for ensuring grain quality and safety, with specific limits on the acceptable levels of contaminants, such as mold, mycotoxins, and insect parts. Other international organizations, such as the Food and Agriculture Organization (FAO) and the World Health Organization (WHO) \cite{FAO_WHO_CA}, have also formulated safety and quality guidelines for grains. The standards may vary depending on the region, and individual buyers or processors may have their own quality standards or specifications. Typically, an inspection of grain kernels is performed through physical testing \cite{rasper2000quality}, sensory evaluation \cite{vithu2016machine}, and chemical analysis \cite{serna2016cereal}. Physical testing and sensory evaluation involve measuring the size, weight, or density of kernels, identifying foreign materials or contaminants, as well as checking for physical signs of damage, discoloration, mold growth, or insect infestation on the surface of the kernel. Chemical analysis is conducted to quantitatively determine precise nutritional value and composition, such as moisture, protein, fat and heavy metal content. These methods are crucial in ensuring the safety and overall quality of the final product by detecting potential contaminants or defects in the grain kernels \cite{patricio2018computer}. 

Generally, the majority of inspections can be determined by observing the surface of grain kernels, termed Grain Appearance Inspection (GAI) in this study. GAI mainly involves two aspects: physical characteristics and Damaged \& Unsound (DU) grains \cite{moss1975sand}. Specifically, physical characteristics pertain to the morphology, size, weight and density of kernels, which are indicators for enhancing processing efficiency and minimizing wastage. DU grains refer to kernels that are damaged or defective, such as moldy, insect-infested, discolored, shriveled, or have heat or moisture damage and physical damage such as cracks or breaks. The proportion of DU grains is one of the essential factors in grain processing, storage and trade. For example, if the percentage of DU grains is less than 2\%, the grain samples can be classified as high-quality cereals, while exceeding 12\% would render them unusable for human consumption and relegated for usage as animal feed or ethanol production.

However, manually conducting GAI is challenging for inspection accuracy and efficiency due to two main aspects. Firstly, GAI requires inspectors to manually take a volume of grain samples (\eg 60 grams (g) and near 1,500 kernels for wheat according to established standards \cite{ISO5527}), and inspect them individually in a kernel-by-kernel way. This is a highly labor-intensive and time-consuming process, as inspectors must concentrate on the surfaces of grain kernels. On the other hand, the surfaces of kernels are heterogeneous and inspectors are required to acquire experience in identifying grain kernels from various regions and species. As a result, there are often variances and discrepancies between inspectors from different expertise levels or regions, particularly for those inspectors working in unique geographical and cereal species conditions. 

In recent years, several studies have addressed GAI-related challenges by leveraging deep learning techniques. For example, Kozlowski \etal \cite{10.1007/978-3-030-23762-2_17} collected near 30\textit{K} barley images, and applied deep convolutional neural networks to perform classification tasks for identifying 4 types of defects. Similarly, Shah \etal \cite{SHAH2022100036} built a dataset including 1432 images of various barley grains, and proposed an efficient deep learning-based technique that achieved 94\% classification accuracy. Furthermore, Fan \etal \cite{fan2023ai4graininsp} introduced the OOD-GrainSet, including 180$K$ and 40$K$ wheat and maize images respectively. It serves as a benchmark for our-of-distribution detection within the realm of grain quality assessment. Velesaca \etal \cite{9150684} developed the CORN-KERNEL dataset, including near 300 pairs of corn images and annotations. The datasets from these studies are typically restricted to sub-species classification tasks for a single cereal species, and the annotations are confined to only visual and classification information.

In this paper, we present a large-scale database aimed at solving GAI challenges and contributing to the development of smart agriculture. Our database, \textbf{GrainSet}, comprises more than 350$K$ individual kernels of four common types of cereal grains, namely wheat, maize, sorghum and rice, sourced from over 20 regions in four countries. For each kernel, \textbf{GrainSet} provides the surface information in high-resolution images captured by our custom-built prototype device, and experts' annotations including morphology, physical size, DU grain category and other sample collection information. Moreover, we conduct experiments and provide baselines to validate the effectiveness and quality of \textbf{GrainSet} by employing a deep learning-based method, obtaining average F1-scores exceeding 94\% across four types of cereal grains. Utilizing our database, we demonstrate the potential for quantitative analysis from a visual perspective, which has applications in improving crop yield, monitoring quality, accelerating processing time and cost, and ensuring food safety.

\section*{Methods}

\begin{figure*}[b]
	
	\begin{center}
		\includegraphics[width=0.98\textwidth]{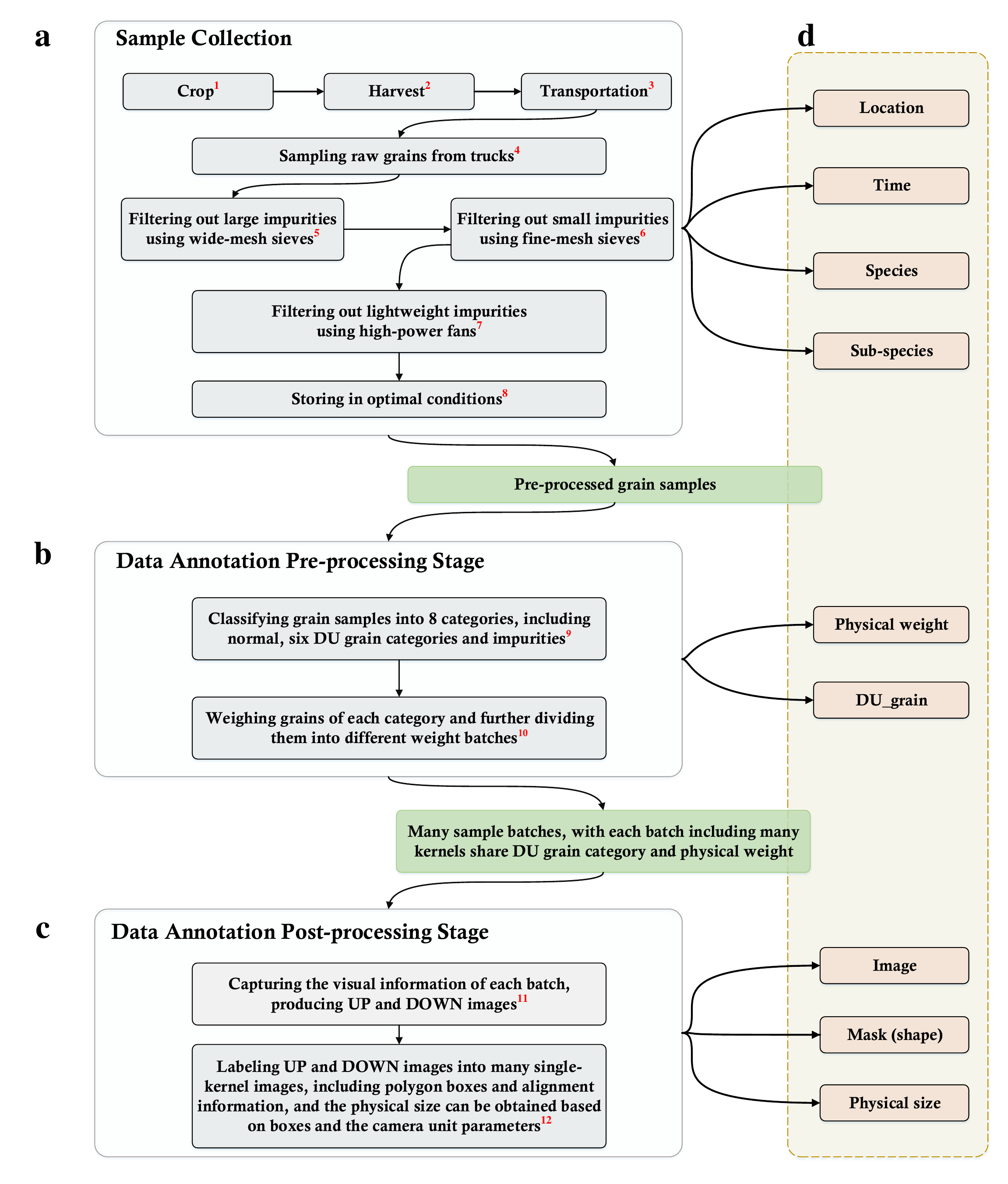}
	\end{center}
	\caption{\textbf{Protocol of data collection and annotations for \textit{GrainSet}}. \textcolor{red}{The red superscript of each step} corresponds to Figure~\ref{fig:whole-pip}. \textbf{a,} Raw cereal grains are sampled from trucks, and undergo a series of processes, including using sieving, high-power fan treatment and storage under optimal conditions. Sample information including location, time, species and sub-species is used as annotations. \textbf{b,} During the pre-processing stage, inspectors classify grain samples into eight categories, including normal, six types of DU grains and impurities. Each category including many samples is further divided into many groups based on physical weight. The physical weight and DU grain information are used as annotations. \textbf{c,} In the post-processing stage, inspectors depict kernel shapes and pair UP and DOWN angles. The images, shapes and physical size are used as annotations. \textbf{d,} Annotations from different stages are stored in the .XML file.}	
	\label{fig:protocol}
\end{figure*}

\begin{figure*}[b]
	
	\begin{center}
		\includegraphics[width=0.98\textwidth]{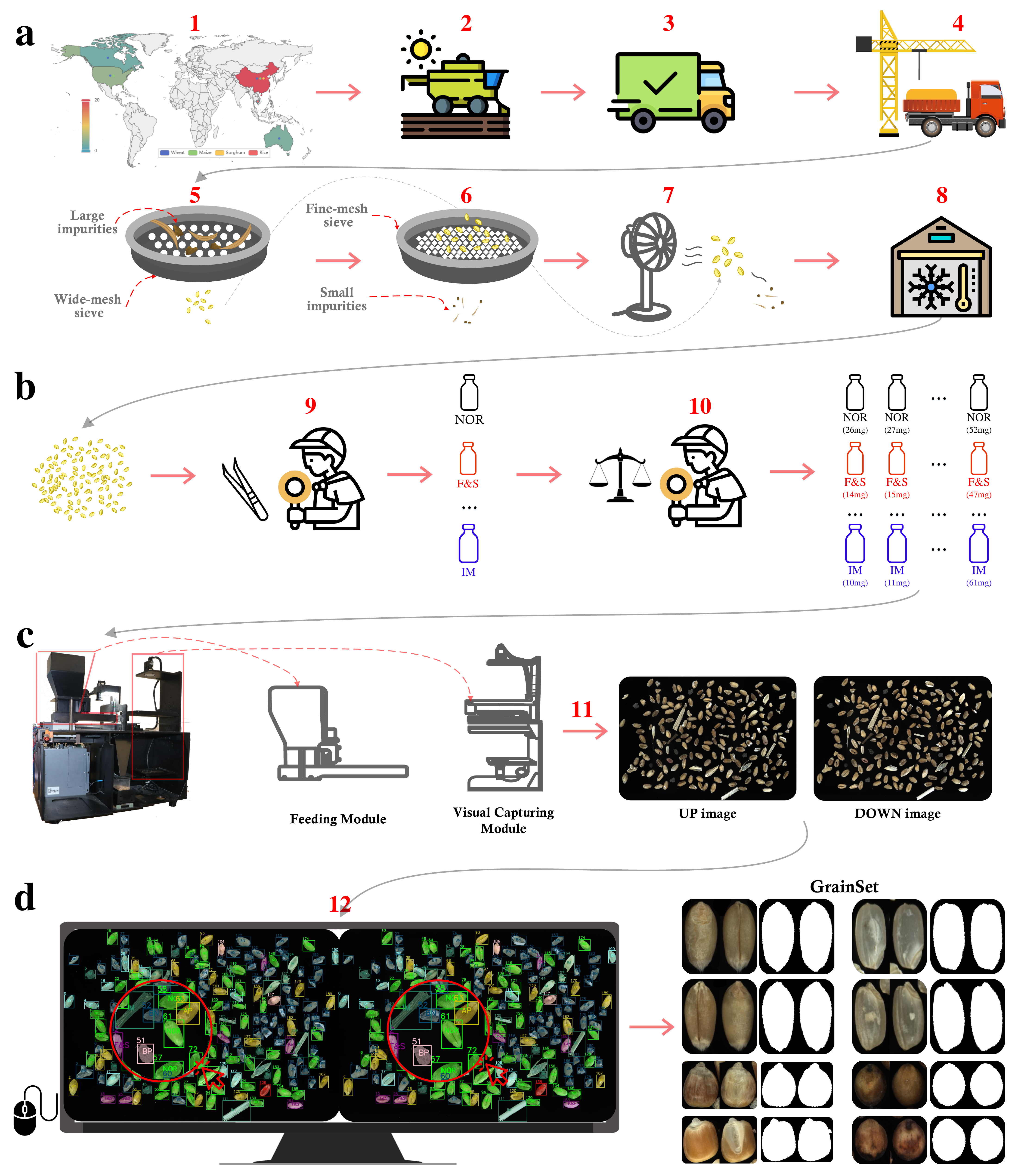}
	\end{center}
	\caption{\textbf{Overview of data collection and annotations for \textit{GrainSet}.} \textcolor{red}{The red superscript of each step} corresponds to Figure~\ref{fig:protocol}. \textbf{a,} \textit{GrainSet} consists of four common types of cereal grains: wheat, maize, sorghum and rice, and raw cereal grains are collected from over 20 regions in 5 countries. Raw samples undergo a series of processes to remove impurities. \textbf{b,} Inspectors divided grain kernels into predefined batches, with each batch sharing the same DU grain category and physical weight information. \textbf{c,}  Our prototype device is built to capture the visual information of each batch, producing a pair of high-resolution UP and DOWN images. \textbf{d,} Inspectors manually depict morphology and align single-kernel images from UP and DOWN angles. All single-kernel images and annotations are combined as \textbf{GrainSet}. }	
	\label{fig:whole-pip}
\end{figure*}

To establish a large-scale and comprehensive database for the purpose of visual quality inspection of grain kernels, we developed a protocol including sample collection, data acquisition and annotations, as shown in Figure~\ref{fig:protocol} and Figure~\ref{fig:whole-pip}. The initial step involves collecting a large volume of cereal grains of diverse characteristics, and then those kernels underwent a series of processes to produce relatively clean samples (see Figure~\ref{fig:protocol}.a and Figure~\ref{fig:whole-pip}.a). 
All grain kernels were manually divided by inspectors into many batches based on their DU category and physical weight information (see Figure~\ref{fig:protocol}.b and Figure~\ref{fig:whole-pip}.b). A prototype device was built to acquire the visual information for grain kernels individually in each batch in a high-efficient way (see Figure~\ref{fig:protocol}.c and Figure~\ref{fig:whole-pip}.c). Then, all captured images labeled by inspectors and corresponding annotations from data collection and the pre-processing stage are combined as \textbf{GrainSet} (see Figure~\ref{fig:protocol}.d and Figure~\ref{fig:whole-pip}.d).

\subsection*{Sample Collection of Cereal Grains}

We collected a high quantity of grain kernels across multiple species and regions between the years of 2016 and 2022. Specifically, the collection effort concentrated primarily on the four most prevalent types of cereal grains, namely wheat, maize, sorghum and rice, which together serve as the majority and comprise more than 90\% of the world grain cereal production according to the report from the Food and Agriculture Organization of the United Nations in 2022 \cite{FAO_reports}. Raw cereal grains were sampled from over 20 regions within 5 countries ranging from Australia (AU), Cambodia (KHM), Canada (CAN), China (CN) and The United States of America (USA), as illustrated in Figure~\ref{fig:whole-pip}.a. We mainly collected wheat for \textit{hexaploid common species} (also named the \textit{bread wheat}) sub-species, maize for \textit{popped corn} and \textit{dent corn} sub-species, sorghum for \textit{sorghum propinquum} sub-species, and rice for \textit{Long}, \textit{Wuchang} and \textit{Jasmine} sub-species.

To ensure the database remains diverse and representative, we collected only a small number of raw cereal grains (with a fixed weight) using skewers from each truck manually, in accordance with sampling guidelines \cite{ISO24333}. The amount of grain sampled is dependent on the physical size and weight of cereal species. Specifically, a laboratory sample of 400$g$ was extracted from a shipment of 30$T$ (tons) for wheat, a sample of 3000$g$ from a shipment of 30$T$ for maize, a sample of 200$g$ from a shipment of 1$T$ for sorghum and a sample of 250$g$ from a shipment of 20$T$ for shelled rice. All those laboratory samples include near 9,000 kernels individually. In summary, we collected a total of 14,000$K$ wheat kernels, 1,200$K$ maize kernels, 6,000$K$ sorghum kernels and 1,500$K$ rice kernels. Those collected raw grains were further pre-processed and filtered.

Subsequently, following the guidelines established by ISO24333 \cite{ISO24333}, the raw grain samples were pre-processed based on a strict protocol, as illustrated in Figure~\ref{fig:protocol}.a and Figure~\ref{fig:whole-pip}.a. The large and small impurities (such as rock or clay) were filtered out using wide-mesh and fine-mesh sieves respectively, and lightweight impurities (such as bran or roots) were removed using high-power fans. It is noted that there exist impurities and extra matter that cannot be filtered through the above pre-processing protocol, since those impurities have similar shapes and physical weights to grain kernels. Those impurities were determined and labeled by inspectors manually in the annotation step. Finally, the pre-processed samples were stored in depositories that provide optimal conditions in temperature and moisture to ensure their preservation. These samples were then subjected to follow-up visual inspection and physical testing.

\subsection*{Image Acquisition Device}

We further built a prototype device to capture the visual information of the surface of grain kernels. To conduct high-efficiency and high-quality acquisition for the four types of cereal grains, there exist two main challenges. First, given that the physical size of various types of cereal grains is typically small, especially the size of sorghum kernels often being smaller than $4$$\times$$4$$\times3 mm^3$, the tiny size imposes a high demand on the precision of the vision system. Second, due to the heterogeneity of kernels in terms of size and shape, kernels are easily pilled up and often occlude each other, which hinders inspection efficiency. 

To address these challenges, we built a machine vision-based prototype device to capture visual information of grain kernels. Our device is able to process a volume of grain kernels simultaneously to achieve a high efficiency of acquisition. It mainly consists of two modules: a feeding module and a visual capturing module, as shown in Figure~\ref{fig:whole-pip}.c. The feeding module was designed to disperse a certain quantity of grain kernels, preventing them from piling up on top of each other. The visual capturing module is then able to capture surface information of those kernels, resulting in efficient and high-quality information acquisition. Specifically, the feeding module is equipped with a vibration belt that delivers pre-processed raw grain kernels, separating them and sequentially transferring them into the visual capturing module. The visual capturing module is equipped with two camera units, and each comprises a high-resolution industrial optical camera with a resolution of $5480$$\times$$3648$ pixels, a focal length of $25mm$ and an object length of $270mm$, enabling the effective field-of-view to reach $131mm \times87mm$ with close to 860 dots-per-inch. Additionally, the visual capturing module also incorporates circle light resources to provide no-shadow light conditions for capturing high-quality images. Two camera units are arranged vertically to capture visual information of kernels from both top and bottom angles, with a transparent plate that is used to hold all kernels and placed in between the two camera units. The surface information of kernels is captured by the camera units when grain kernels are stationary after being transformed from the feeding module. Due to the heterogeneity of kernel shapes and the presence of concave corners, a smaller region (less than $2\%$) of the surface may not be captured. In summary, the prototype device can acquire visual information for a batch of grain kernels simultaneously, producing two high-resolution images named UP and DOWN images corresponding to the two camera units. 

\subsection*{Image Annotations}

We built a data annotation protocol to conveniently provide annotation information for grain samples. The protocol consists of pre-processing and post-processing stages. The pre-processing stage is conducted on raw grain samples, as shown in Figure~\ref{fig:protocol}.b and Figure~\ref{fig:whole-pip}.b. Inspectors manually divided kernels into eight categories including normal, six DU grain categories and impurities (which will be introduced in the next section). Then, samples in each category were further divided into multiple batches based on their physical weights, with a weight interval of 1$mg$ for each batch. It is noted that the DU grain and physical weight information was annotated by expert inspectors in a laboratory environment equipped with specialized tools (\eg tweezers and scales), and each annotation was confirmed by at least two independent inspectors through blind validation. As a result, a batch of kernels share the same species, physical weight and DU grain category, and such information serves as the annotations directly when our device captures visual information and produces a pair of UP and DOWN images for this batch.

The post-processing stage was conducted on UP and DOWN images, as shown in Figure~\ref{fig:protocol}.c and Figure~\ref{fig:whole-pip}.d. To be specific, inspectors manually annotated every kernel in high-resolution images, and each kernel was labeled with a pixel-level polygon box to depict the morphological shape, and then physical size can be calculated according to the shape and fixed field-of-view of the camera units. Subsequently, each high-resolution image can produce a number of single-kernel images based on annotated polygon boxes. For both UP and DOWN images, two sets of polygon boxes can be obtained and are used to localize and crop the visual information of every single kernel from two different angles. To obtain comprehensive visual information for each grain kernel, a straightforward approach is to pair the polygon boxes based on the distance between their centroids. However, in practice, we observed that the kernels in UP and DOWN images may not align perfectly due to inherent machine limitations, and the kernels may shift slightly when transferred from the feeding module to the visual capturing module in some cases. Hence, inspectors manually matched two single-kernel images from both UP and DOWN images. As a result, our \textbf{GrainSet} database contains many single-kernel images and corresponding data collection and annotation information.

\section*{Data Records}
\label{data}

The \textbf{GrainSet} is deposited on the Figshare under the Creative Commons BY 4.0 license. The whole database includes both processed single-kernel images and annotations. The \textbf{GrainSet} is comprised of four sub-datasets corresponding to different grain types, \ie wheat, maize, sorghum and rice, with each sub-dataset containing numerous single-kernel images with corresponding annotations. All images are stored in the Portable Network Graphics (PNG) format due to its lossless compression capabilities, in which color information is encoded using the RGB (Red, Green, Blue) color model.
It is noted that we zip each sub-dataset as a single file. Additionally, considering that the complete \textbf{GrainSet} is large in disk storage, we randomly select near 2\% samples from \textbf{GrainSet} to construct a preview dataset named \textbf{GrainSet-tiny} for a preview of our \textbf{GrainSet}. We additionally release 5\% samples of the raw images captured by our acquisition device to create \textbf{GrainSet-raw}, including high-resolution images and comprehensive annotations (\eg kernel localization information, DU categories). It serves as a reference for understanding the data acquisition and pre-processing procedures.

\subsection*{Statistics of \textit{GrainSet}}

\begin{figure*}[b]
	
	\begin{center}
		\includegraphics[width=1\textwidth]{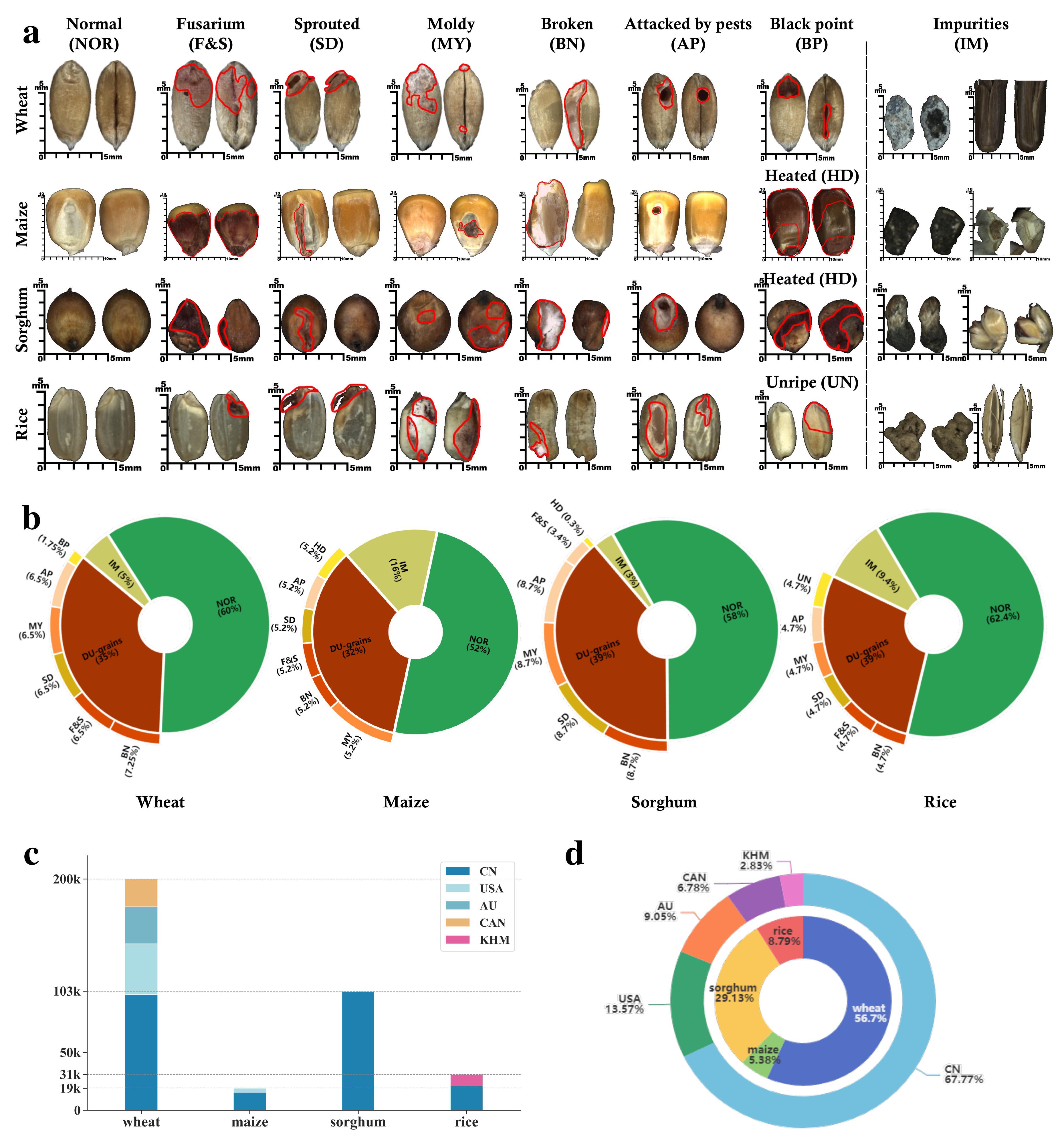}
	\end{center}
	\caption{\textbf{Examples and distributions of sample information in \textit{GrainSet}.} \textbf{a,} Examples of normal, each kind of DU grains, and impurities for four types of cereal grains. The abbreviations (\eg NOR) are used in subsequent content. Red lines highlight discriminative regions. \textbf{b,} Percentages of normal, each kind of DU grains, and impurities for four types of cereal grains. \textbf{c,} Percentages and numbers of regional information for four types of cereal grains. \textbf{d,} The inner pie: percentages of four types of cereal grains in \textbf{GrainSet}. The outer pie: percentages of regional information in \textbf{GrainSet}.}	
	\label{fig:data-stat}
\end{figure*}

The overview of \textbf{GrainSet} is presented in Table \ref{tab:stats}. It consists of a total of 352.7$K$ single-kernel images with 200$K$, 19$K$, 102.7$K$ and 31$K$ images for wheat, maize, sorghum and rice, respectively. For wheat, we selected 200$K$ wheat kernels from 14,000$K$ raw grain samples that were collected from over 10 regions in 4 countries. To construct a comprehensive database, we maintained the distribution balance in terms of DU grain category and regions, with detailed distributions shown in Figure~\ref{fig:data-stat}. However, collecting a significant number of different kinds of DU grain kernels is challenging, as the proportion of DU grains is small. We tried our best to maintain the balance of data distribution in terms of the DU grain category. The normal and DU grains account for 60\% and 35\% of the total respectively, and the ratio of impurities that are not filtered out through the pre-processing protocol accounts for 5\%. 

Similar to wheat, we selected 19$K$ maize kernels from near 1,200$K$ raw grain samples that were collected from over 6 regions in 2 countries including China and The United States of America. The normal, DU grains and impurities account for near 52\%, 32\% and 16\% of the total respectively. For sorghum, we selected near 102$K$ sorghum kernels from more than 6,000$K$ raw grain kernels that were collected from over 3 regions in China. The normal, DU grains and impurities account for near 58\%, 38\% and 4\% respectively, since the ratio of impurities in raw samples is rare. For shelled rice, we selected 31$K$ rice kernels from 1,500$K$ raw grain samples that were collected from over 4 regions in 2 countries including China and  Cambodia. The normal, DU grains and impurities account for near 65\%, 29\% and 6\% respectively. Additionally, we are still updating the database by enriching more grain types and expanding the number of grain samples.

\subsection*{Normal, DU grains and Impurities}

Each type of cereal grain can be classified into eight categories according to ISO5527 \cite{ISO5527}. Figure~\ref{fig:data-stat}.a illustrates representative examples of each of the eight categories. 
For all types of cereal grains, the normal (NOR) grain refers to kernels that have not undergone any sprouting process or other damage. These grains are commonly consumed as staple foods or used in various food products. Impurities (IM) refer to extra materials that cannot be filtered through the data collection protocol.

Among DU grains, the Fusarium \& Shriveled (abbreviated F\&S) grain indicates kernels infected by fungi belonging to the Fusarium genus. These grains produce toxic substances and have negative effects on safety. The sprouted grain (SD) refers to kernels that have begun the process of germination or sprouting. The moldy grain (MY) refers to grains contaminated with mold or fungi. The broken grain (BN) indicates grains that have been fractured or fragmented during harvesting or transportation. The grain attacked by pests (AP) refers to grains that have been infested by insects. The black point grain (BP) indicates grains (particularly wheat in this study) affected by factors including fungal infections or insect damage. The heated grain (HD) refers to grains (particularly maize and sorghum) that have undergone an unintentional or excessive increase in temperature during storage or processing. The unripe grain (UN) indicates grains (particularly rice) that have not fully reached their maturity or optimal ripeness before being harvested. Among these DU grains, F\&S, MY and BP grains have negative effects on health and safety, while SD, BN, AP, HD and UN grains have decreased value.

\subsection*{Annotation Files}

Each sub-dataset is accompanied by an Extensible Markup Language(.XML) file containing a large number of elements, with an element representing an individual grain kernel and containing a variety of attributes in terms of sample collection and experts' annotation, as illustrated in Table~\ref{tab:xml_des}. The \textit{ID} attribute refers to the unique identifier enabling retrieval of the single-kernel image and the pixel-level binary mask that depicts the shape of the kernel, and all masks are stored in PNG format. The attributes \textit{species}, \textit{sub-species}, \textit{location} and \textit{time} provide information pertaining to the grain during the sample collection. To avoid potential ethical issues or privacy concerns, any specific region information is erased and location information is presented only at the country level. For experts' annotation, \textit{size} is calculated based on the annotated mask. The attributes \textit{DU\_grain} and \textit{weight} are annotated by expert inspectors during the data annotation pre-processing stage.

\section*{Technical Validation}

\textbf{GrainSet} is constructed for GAI-related tasks, and multiple real-world GAI applications can be formulated and conducted based on \textbf{GrainSet}. For example, counting the exact number of kernels in a laboratory sample, classifying the types of cereal grains and impurities, and recognizing the DU grains or sub-species. To validate our \textbf{GrainSet}, we conducted classification tasks for recognizing DU grains for all four types of cereal grains, and this task is to classify samples into eight categories including normal, six types of DU grains and impurities.

For each type of cereal species, we randomly split each sub-dataset into training, validation and test sets, with 80\%, 10\% and 10\% proportions respectively by preserving the ratios of normal, each of DU grain categories and impurities, and we denoted such settings as the \textit{whole}. On the other hand, considering the imbalanced distributions within our database, we also built a \textit{balanced} sub-set as another training set, in which each category comprises an equal number of images based on the minimum number of images among all DU categories.

\begin{figure*}[b]
	\begin{center}
		\includegraphics[width=0.9\textwidth]{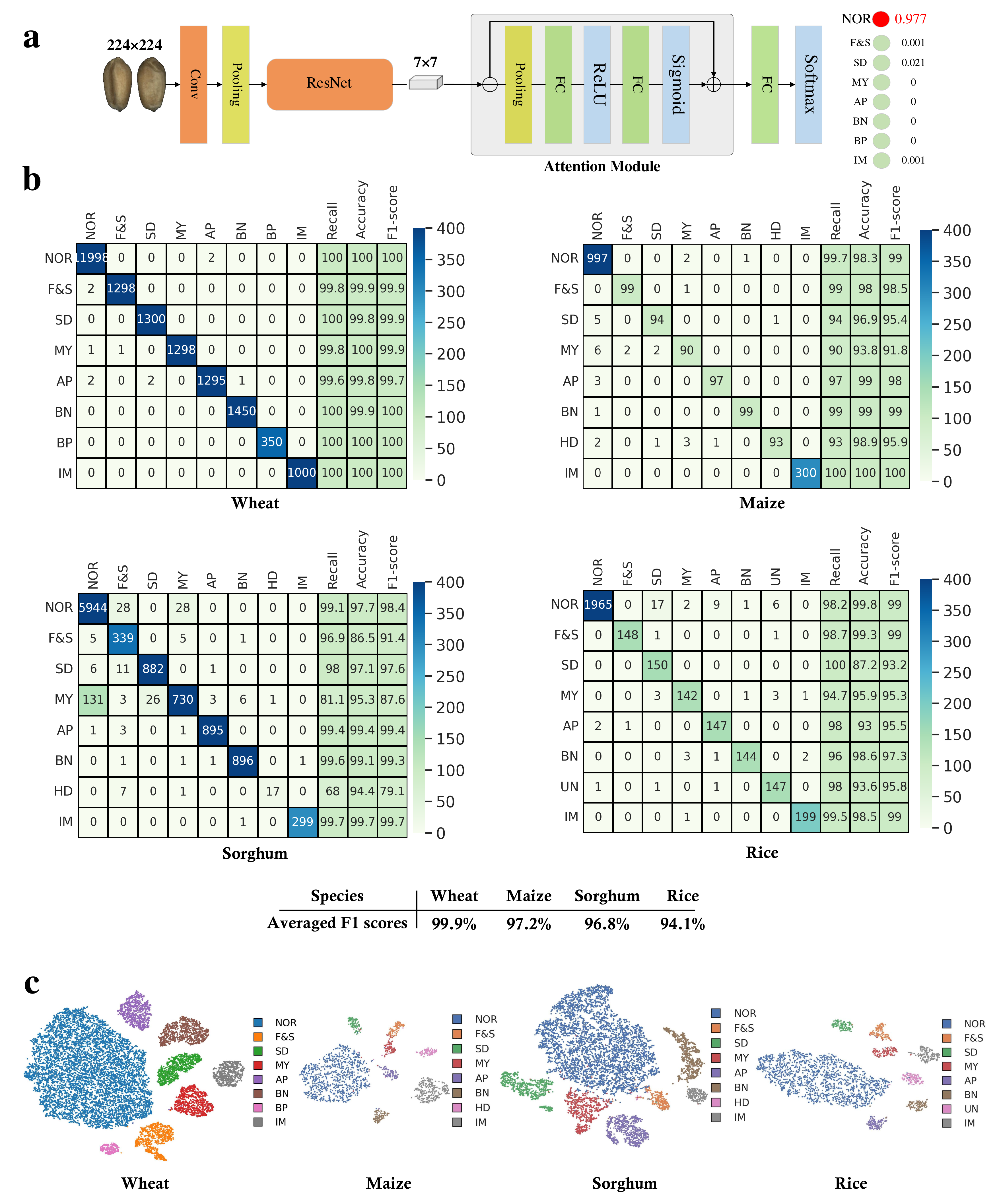}
	\end{center}
	\caption{\textbf{Overview of the model and results of technical validation.} \textbf{a,} The detailed structure of our fine-grained classification model. It employs ResNet-50 as the backbone and SE-attention module as the prediction head. \textbf{b,} The performance of our trained models on test sets for four types of cereal grains. The recall, accuracy and F1-score are reported in the confusion matrix \textbf{c,} The t-SNE visualization results of features on test sets for four types of cereal grains.}	
	\label{fig:models}
\end{figure*}

To build our classification models, we employed advanced deep convolutional neural networks (CNNs) \cite{he2016deep}, which have achieved remarkable progress in multiple fields including medical image analysis \cite{9980424}, industrial applications \cite{bergmann2019mvtec}, and agriculture \cite{fan2022grainspace}. In this work, we use a widely-used classification model to assess the quality of our database. Our model consists of a backbone and an attention head module, as depicted in Figure~\ref{fig:models}.a. The backbone employed the classical ResNet-50 \cite{he2016deep} which includes consecutive residual convolutional block and down-sampling convolutional layers to extract semantic features from input images. The attention head module employed a Squeeze-and-Excitation (SE) module \cite{hu2018squeeze}. The SE module consists of channel-wise global pooling operations, fully convolutional layers, and ReLU and sigmoid activation layers, which can adaptively re-calibrate channel-wise feature responses to focus on discriminative regions. Following the success of transfer learning techniques \cite{zhuang2020comprehensive}, we trained four models for four types of cereal grains respectively, and initialized all parameters of backbones by using the models pre-trained on ImageNet \cite{deng2009imagenet}. Then, to fine-tune each classification model, we trained it on a GPU workstation with the PyTorch platform \cite{paszke2019pytorch}. The training epoch and batch size were set to 50 and 128. A step learning strategy was employed and an initial learning rate was set to 0.0012, and an SGD optimizer was employed with a weight decay of 0.0001. All input images were resized into $224$$\times$$224$ pixels, and data augmentations including random flips, rotations and color jitters were applied. CrossEntropy loss was used as the optimization objective. It is noted that we trained four classification models for four kinds of cereal grains respectively with similar training parameters. Additionally, we conducted experiments to assess the performance of classical and deep network-based methods on our proposed database. For classical methods, we employed color histogram and Scale-Invariant Feature Transform (SIFT) \cite{lowe2004distinctive} for extracting color and texture information, and utilized Support Vector Machine (SVM) \cite{708428} for classification. For deep networks, we evaluated various network architectures, including VGG19 \cite{simonyan2014very}, Inception-v3 \cite{szegedy2016rethinking}, ResNet-50 \cite{he2016deep}, ResNet-152 \cite{he2016deep} and Vision Transformer (ViT) \cite{dosovitskiy2020image}. All training parameters are similar to our model.

\begin{figure*}[b]
	\begin{center}
		\includegraphics[width=0.9\textwidth]{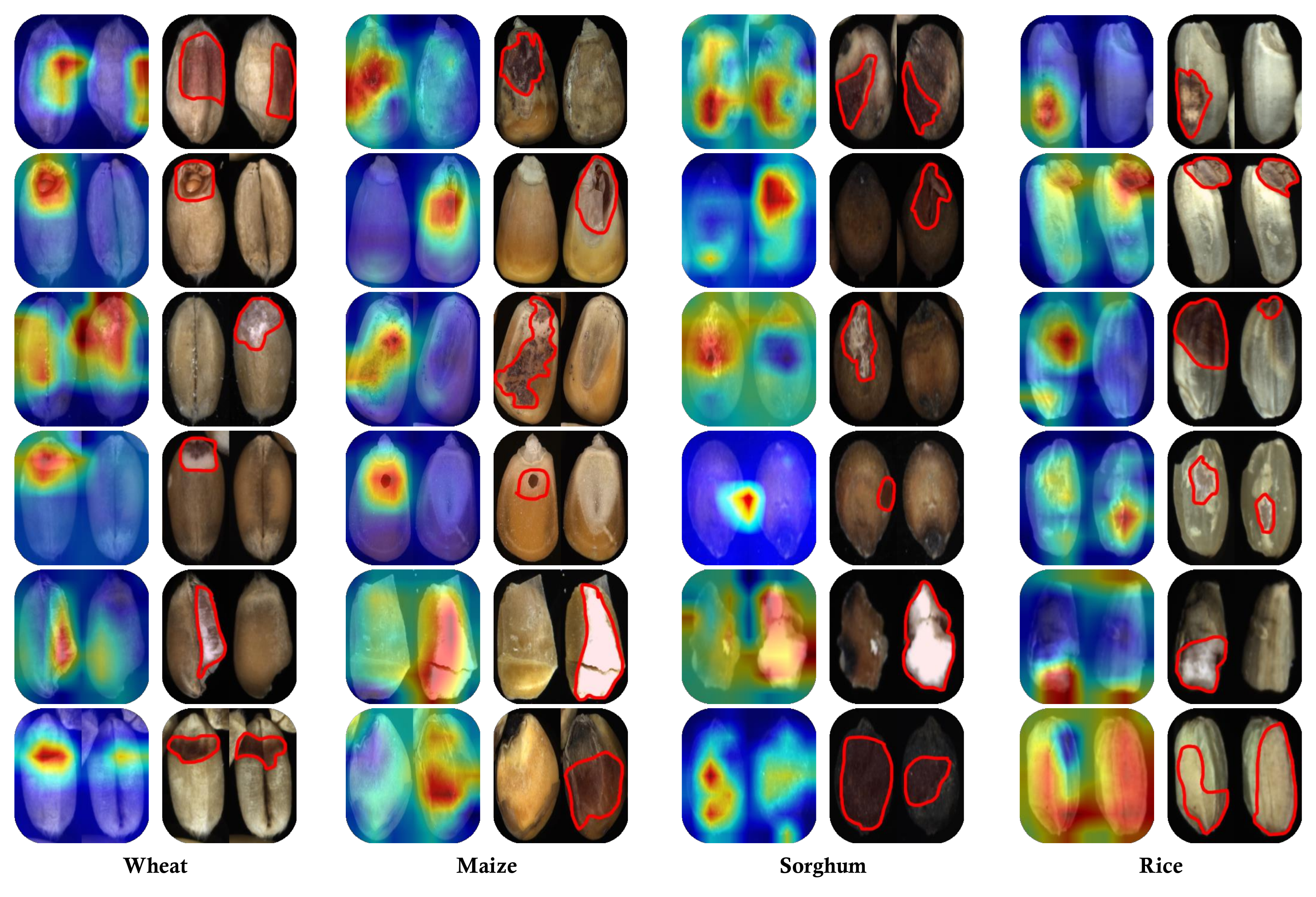}
	\end{center}
	\caption{\textbf{The Gard-CAM visualization results extracted from our trained models, and red lines highlight the discriminative regions.}}	
	\label{fig:cam}
\end{figure*}

The performance comparison of classical methods, various network backbones and our model under both \textit{whole} and \textit{balanced} settings is presented in Table~\ref{tab:backbones}. It can be seen that employing the \textit{whole} setting achieves better performance compared to the counterparts using the \textit{balanced} setting. This is because there are many more training samples with the \textit{whole} setting, which can substantially benefit feature learning. Given the extreme imbalance between normal and DU cereal grains in actual scenarios, we recommend that future work approach the issue from the perspective of addressing this imbalance. Using color histograms and SIFT as feature extractors yield 70.6\% and 35.5\% macro F1-scores on the \textit{whole} setting respectively. Deep networks-based methods show obvious improvements, in which ResNet-152 achieves the best performance of 96.4\% and 96.1\% average accuracy and macro F1-score respectively. Compared to these methods, the detailed performance of our models is illustrated in Figure.~\ref{fig:models}.b, where we show the confusion matrix, recall, accuracy and F1-score together. It can be observed that our models produced remarkable classification results, achieving average F1-scores of 99.9\% for wheat, 97.2\% for maize, 96.8\% for sorghum and 94.1\% for rice respectively. We also visualized qualitative results to further validate our results, as shown in Figure~\ref{fig:models}.c and Figure~\ref{fig:cam}. Specifically, we employed the t-SNE technique \cite{van2008visualizing} to visualize the intermediate features that are compressed and extracted from the output of the SE attention head. We observed that normal, different kinds of DU grain categories and impurities are distinctly separated in the feature space, verifying the quality of our \textbf{GrainSet}. We also employed the Gard-CAM technique \cite{selvaraju2017grad} to visualize the activated areas where our trained models focus on the input images. It can be observed that our model can effectively focus on the discriminate regions with distinct characteristics of DU grains. For example, our models can concentrate on moldy spots, wormholes and broken parts for four types of cereal grains. Overall, our experimental results demonstrate the quality of our \textbf{GrainSet}.

\section*{Code availability}

The validation code and models are released in the Github repository \url{https://github.com/GrainSpace/GrainSet}. We conducted benchmark analysis using PyTorch and OpenCV, and all parameters used in our implementation are reflected in the repository.

\newpage
\section*{Figures \& Tables}

\begin{table}[h]
  \centering
    \resizebox{0.98\textwidth}{!}{
    \begin{tabular}{cccc|cccc}
    \toprule
    \multirow{2}[0]{*}{Species}  & \multicolumn{4}{c}{Kernel Information} & \multirow{2}[0]{*}{Regions} & \multirow{2}[0]{*}{ Disk storage} & \multirow{2}[0]{*}{Figshare DOI}\\
    \cmidrule{2-5}
      & Normal & DU grains & Impurities & Total & &  & \\
    \midrule
    Wheat &  120$K$ & 70$K$ & 10$K$ & 200$K$ & CN, CAN, AU, USA  & 19.6$G$ & \url{https://doi.org/10.6084/m9.figshare.22992317.v2} \\
    Maize  & 10$K$ & 6$K$ & 3$K$ & 19$K$ & CN, USA & 5.6$G$ &\url{https://doi.org/10.6084/m9.figshare.22987562.v2} \\
    Sorghum &  60$K$ & 39.7$K$ & 3$K$ & 102.7$K$ & CN  & 6.9$G$ &\url{https://doi.org/10.6084/m9.figshare.22988981.v2} \\
    Rice  & 20$K$ & 9$K$ & 2$K$ & 31$K$ & CN, KHM  & 2.4$G$ & \url{https://doi.org/10.6084/m9.figshare.22987292.v3}\\
    \cmidrule{1-8}
    \multicolumn{4}{c}{\textbf{GrainSet-tiny}} & 6.5$K$ & - & 0.6$G$ & \url{https://doi.org/10.6084/m9.figshare.22989029.v1}\\
    \multicolumn{4}{c}{\textbf{GrainSet-raw}} & 15$K$ & - & \textcolor{blue}{3.2$G$} & \url{https://doi.org/10.6084/m9.figshare.24137472.v1}\\
    \bottomrule
    \end{tabular}%
    }
    \caption{\textbf{The statistical information of \textit{GrainSet}}. \textbf{GrainSet} consists of four sub-datasets corresponding to four types of cereal grains, including wheat, maize, sorghum and rice. We randomly select 2\% samples from \textbf{GrainSet} to construct \textbf{GrainSet-tiny}, which is a preview for understanding our database. We randomly select 5\% raw images captured by our acquisition device to create GrainSet-raw, which serves as a reference for understanding the data acquisition and pre-processing procedures.}
    \label{tab:stats}%
\end{table}

\begin{table}[h]
  \centering
    \resizebox{0.98\textwidth}{!}{
    \begin{tabular}{cccc}
    \toprule
    \multicolumn{2}{c}{Attribute}  & Descriptor & Data field (example)  \\
    \midrule
    \multicolumn{2}{c}{ID}  & \makecell[c]{The unique identification of a grain kernel, with the format as \\ \textit{Grainset\_<species>\_<sampling time>\_<IDinImage>\_<DeviceID>}}
      &  Grainset\_wheat\_2022-11-23-13-42-01\_32\_p600s  \\
    \cmidrule{1-4}
    \multirow{4}[0]{*}{\makecell[c]{Data \\ collection} } &  species & The species of the cereal grain, one of (\textit{wheat, maize, sorghum or rice }) & wheat   \\
    & sub-species & The sub-species of the grains, \eg one of (\textit{Long, Wuchang, Jasmine}) for rice  & hexaploid common \\
    &  location & The country information where the kernel is sampled & CN  \\
    &  time & The time when the kernel is sampled, with the format as \textit{<year-mm-dd>} & 2019-11-10\\ 
    \cmidrule{1-4}
    \multirow{3}[0]{*}{Annotation} &  size &  The physical size of the kernel and the unit is mm$^2$ &  19\\
    &  DU grain & The DU grain category of the kernel & NOR \\
    &  weight & The physical weight of the kernel and the unit is gram & 30  \\
    \bottomrule
    \end{tabular}%
    }
\caption{\textbf{The annotation file is in .XML format}. A sub-dataset includes a .XML file with a large number of elements, indicating eight attributes including ID, species, sub-species, location, time, size, DU grain and weight. These attributes are annotated from two stages: sample collection and experts' annotation.}
  \label{tab:xml_des}%
\end{table}

\begin{table}[h] 
  \centering
    \resizebox{0.88\textwidth}{!}{
    \begin{tabular}{cccccc}
    \toprule
    \multicolumn{2}{c}{\multirow{2}[0]{*}{Methods}} & \multicolumn{4}{c}{\textit{whole} / \textit{balanced} settings (Accuracy, F1-scores \%)} \\
    \cmidrule{3-6}
    \multicolumn{2}{c}{} & Wheat & Maize & Sorghum & Rice \\
    \midrule
    \multirow{2}[0]{*}{Classical} & Color & 74.8/72.4, 71.8/75.4 & 82.7/74.5, 78.5/72.2 & 81.3/69.2, 67.6/69.9 & 64.7/55.2, 64.7/59.3 \\
          & Texture & 42.2/30.9, 31.6/32.5 & 49.5/40.7, 32.9/40.4 & 53.1/40.5, 41.6/41.8 & 41.9/36.8, 35.8/39.7 \\
    \midrule
    \multirow{5}[0]{*}{Deep Networks} & VGG19 & 98.8/97.2, 98.4/98.0,  & 94.9/88.6, 93.8/89.9 & 97.1/92.9, 94.6/92.5 & 89.6/80.0, 91.5/85.0 \\
          & Inception-v3 & 99.8/98.9, 99.8/99.0 & 96.1/88.2, 95.1/89.6 & 97.2/94.7, 94.9/93.2 & 89.4/87.4, 91.6/90.6 \\
          & ResNet-50 & 99.9/99.3, 99.8/99.3 & 97.0/86.1, 96.1/88.0 & 97.1/94.3, 93.8/90.9 & 89.1/84.0, 92.1/86.9\\
          & ResNet-152 & 99.8/99.1, 99.7/99.3 & 96.7/93.7, 95.7/93.1 & 96.6/92.5, 94.6/88.4 & 92.2/86.8, 94.3/90.6 \\
          & ViT   & 75.9/81.4, 68.1/76.5 & 85.6/80.7, 82.4/81.5 & 66.7/81.8, 65.0/74.1 & 78.1/75.7, 79.4/79.5 \\
          & Our   & 99.9/99.4, 99.9/99.4 & 97.9/94.1, 97.2/93.8 & 96.1/94.0, 96.8/91.6 & 95.7/91.9, 94.1/90.9 \\
    \bottomrule
    \end{tabular}%
    }
\caption{\textbf{Performance of both classical and deep networks-based methods using \textit{whole} and \textit{balanced} settings respectively.} We employed two kinds of classical methods including color (color histogram + SVM) and texture (SIFT + SVM), five kinds of network architectures including VGG19, Inception-v3, ResNet-50, ResNet-152 and ViT, and our model that is based on ResNet-50 and SE module.}
  \label{tab:backbones}%
\end{table}

\clearpage
{
\bibliography{sample}
}

\section*{Author contributions statement}

L.F and Y.S conceived and led the expedition. L.F, Y.D, D.F, Y.W developed data protocols, implemented the image analysis methods, and collected the raw data. M.P provided further technical and scientific advice to the project. L.F, Y.D and Y.S drafted the manuscript, and all authors reviewed and agreed to the published version of the manuscript.

\section*{Competing interests}

The authors declare no competing interests.

\end{document}